\documentclass{article}
\usepackage[british]{babel}
\usepackage{jfrExamplee}
\usepackage{graphicx}
 \usepackage{multirow}
\usepackage{setspace}
\usepackage{subcaption}

\usepackage[style=apa,backend=biber,bibencoding=utf8]{biblatex}
\DeclareLanguageMapping{british}{british-apa}
\AtEveryBibitem{\clearfield{month}}
\AtEveryBibitem{\clearfield{day}}

\usepackage{gensymb}
\usepackage{amsmath}
\usepackage{xcolor}
\usepackage{multicol}
\usepackage{caption}

\usepackage[a4paper, footskip=2cm, top=2.5cm, bottom=3.3cm]{geometry}

\usepackage{lineno}
\title{Dataset and Performance Comparison of Deep Learning Architectures for Plum Detection and Robotic Harvesting}

\author{
Jasper Brown\thanks{Corresponding author}\\
The Australian Center for Field Robotics \\
The University of Sydney \\
Sydney, Australia \\
\texttt{j.brown@acfr.usyd.edu.au} \\
\And
Salah Sukkarieh \\
The Australian Center for Field Robotics \\
The University of Sydney \\
Sydney, Australia \\
\texttt{salah@acfr.usyd.edu.au}
}

\begin{document}
\maketitle

%\subsubsection*{Acknowledgements}

\begin{abstract}

Many automated operations in agriculture, such as weeding and plant counting, require robust and accurate object detectors. Robotic fruit harvesting is one of these, and is an important technology to address the increasing labour shortages and uncertainty suffered by tree crop growers. An eye-in-hand sensing setup is commonly used in harvesting systems and provides benefits to sensing accuracy and flexibility. However, as the hand and camera move from viewing the entire trellis to picking a specific fruit, large changes in lighting, colour, obscuration and exposure occur. Object detection algorithms used in harvesting should be robust to these challenges, but few datasets for assessing this currently exist. In this work, two new datasets are gathered during day and night operation of an actual robotic plum harvesting system. A range of current generation deep learning object detectors are benchmarked against these. Additionally, two methods for fusing depth and image information are tested for their impact on detector performance. Significant differences between day and night accuracy of different detectors is found, transfer learning is identified as essential in all cases, and depth information fusion is assessed as only marginally effective. The dataset and benchmark models are made available online. 


\end{abstract}

% \begin{multicols}{2}

\section{Introduction}
Automated tree crop harvesting is a well known focus in agricultural robotics and has the potential to reduce the high labour costs associated with fruit picking. Accurate target detection is the first step for most harvesting systems and inaccuracies at this early stage can disproportionately affect system performance. Optimising detector accuracy should therefore be a priority in agricultural robotics and automation. As deep learning methods from computer vision have advanced rapidly, the application of these in agriculture has become common place. However, these algorithms require a large and representative corpus of training data to perform well, something not currently available for the full range of image conditions seen in harvesting. 

Eye-in-hand sensing provides distinct advantages for fruit harvesting by allowing for continuous feedback control right up to the point of picking. This decreases positioning error upon gripper final approach. Off-hand sensors provide a much more regular view of target fruit, but are difficult to re-position for better viewing angles. Unfortunately, the added sensor movement of eye-in-hand systems results in much more variable imaging conditions. Easily accessing the tree canopy also dictates the use of a compact, and preferably low cost, sensor. All of these present additional challenges for object detectors deployed in harvesting systems. 

\begin{figure}
	\centering
	\includegraphics[width=0.6\columnwidth]{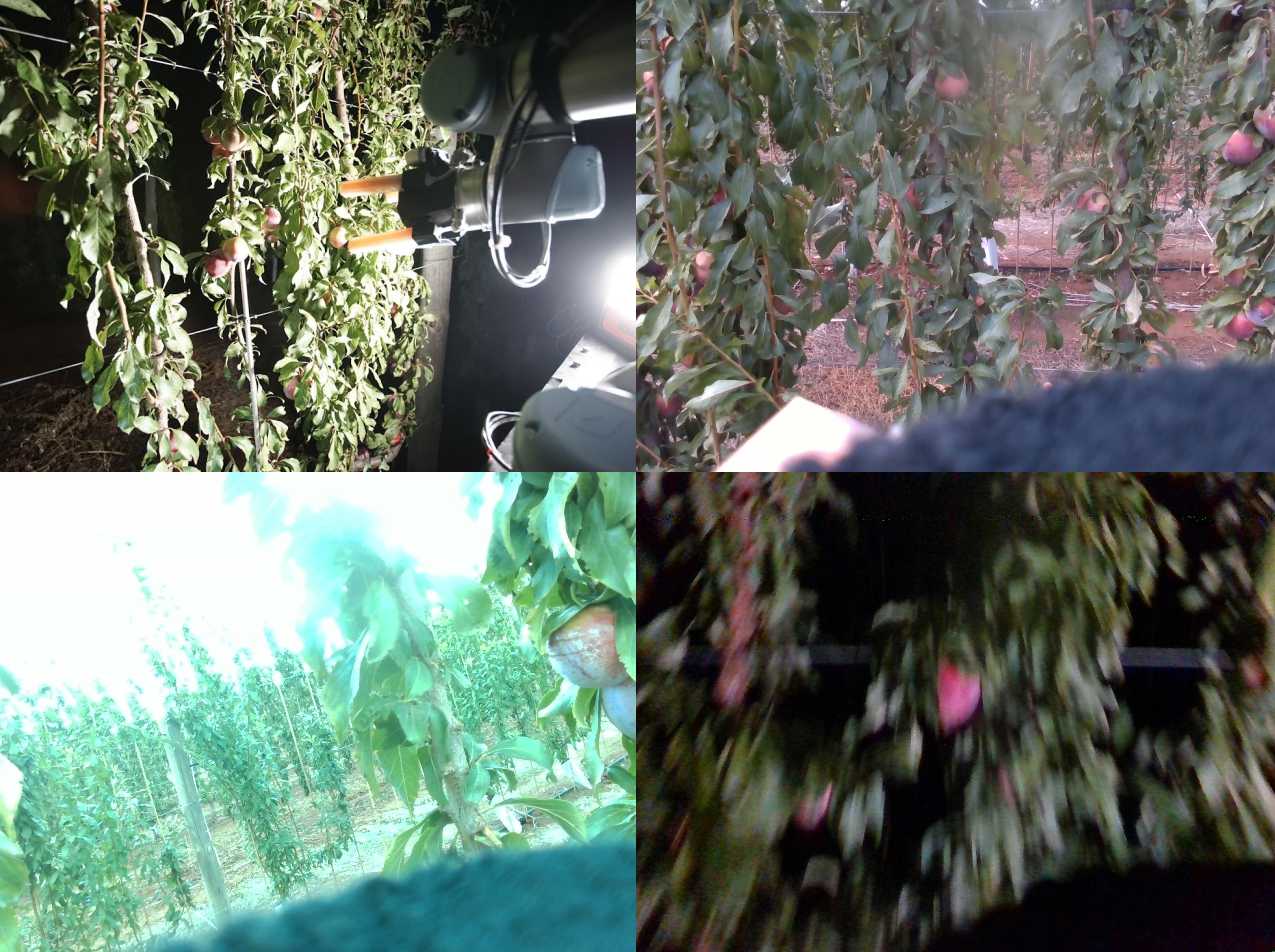}
	\captionof{figure}{Clockwise from top left are examples of; a typical night time eye-off-hand viewpoint, a typical day time eye-in-hand viewpoint, exposure and colour changes due to the camera entering the sun after being obscured by leaves, blurring due to low light levels at night.}
	\label{fig:DataExample}
\end{figure}

To address these challenges, new datasets representative of eye-in-hand harvesting conditions are required. For this, images were gathered during a field evaluation of a prototype system that uses an eye-in-hand configuration for harvesting plums on a 2D trellis~\parencite{brown2020}. A small and cheap Intel Realsense D435 RGB-Depth (RGBD) camera is used. Many image artefacts caused by the eye-in-hand setup can be observed in the dataset, see Figure~\ref{fig:DataExample}. A total of 700 images are extracted and annotated, these are split evenly between day and night datasets. Two previous generation object detection deep learning networks, Faster-RCNN and YoloV3, are benchmarked against this dataset, along with two current networks; RetinaNet and CenterNet. 

% \end{multicols}
% \begin{multicols}{2}

Depth sensing is a required modality for harvesting, so the fusion of depth information for detection is also tested. Both early and late data fusion is trialled using the RetinaNet architecture. Early fusion treats the network input as a 4 dimensional RGBD image, while in late fusion the image and depth features are concatenated after being extracted by parallel network backbones. 

Key contributions of this work are the presentation of two realistic datasets from an eye-in-hand harvesting operation, the benchmarking of multiple current deep learning architectures against these and the assessment of RGBD data fusion for detection. All data and models are available publicly at the link in Section~\ref{sec:dataset}. 

The primary limitation of this work is the relatively small dataset size, which makes it impossible to test which benchmark networks perform best with very large datasets. Numerous works have shown a clear correlation between training set size and network performance for fruit detection, though the exact performance-by-size function is not linear and varies between architectures. 

\section{Related Work}
Fruit detection is a key problem that is common in the literature for goals such as yield estimation, crop health assessment and harvesting~\parencite{koirala2019a,koirala2019b,arad2019,fernandez2018,sa2017,bargoti2017a,bargoti2017b,stein2016}. The most basic form consists of placing bounding boxes around each of the fruit in an image. Traditional computer vision methods have been extensively employed, while many recent works make use of deep learning tools. These require large amounts of training data, though this requirement can be relaxed using data augmentation and simulation methods, or transfer learning techniques. 

Specular reflections from round fruit, combined with local image gradients are used in~\cite{wang2013} for object detection under controlled lighting. A Support Vector Machine (SVM) is trained to detect apples using thermal imagery in~\cite{feng2019}. The accuracy performance of SVM object detectors has now been surpassed by deep learning approaches, though they remain competitive for classification when using low dimensional hand-engineered image features~\parencite{kamilaris2018, gongal2015}. Edge features are used with Hough voting and an SVM classifier by~\cite{sengupta2014} to identify green citrus fruit under varying illumination. Similarly, \cite{maldonado2016} make use of a bas-relief representation with edges, Hough voting and an SVM to also count green citrus fruit. \cite{nguyen2016} perform RGB and depth channel thresholding to identify point cloud blobs corresponding to apples. These are separated using a Euclidean distance metric which is tested in the field under semi-controlled lighting conditions. 

Avocados, apples and lemons are counted using a hand held camera in~\cite{vasconez2020}. Both Faster-RCNN and the Single Shot multibox Detector (SSD) network are tested with a video based object tracker to prevent multiple-counting. Avocados were the most accurately detected fruit for both networks, followed by lemons and apples. This is counter-intuitive given the strong colour features present for the latter two fruit, but may be explained by a larger set of training images for avocados.

Harvesting-oriented apple detection data is gathered in~\cite{kang2020b}. The YoloV3, Mask-RCNN and Faster-RCNN architectures are thoroughly tested, along with their own deep learning model, described in~\cite{kang2019}. This implements the idea of focal loss, similar to RetinaNet. Unfortunately the dataset and trained models are not released for comparison and, unlike the present work, the harvesting system does not use an eye-in-hand camera. \cite{gao2020} train a multi class Faster-RCNN detector to distinguish between different obscuration cases for apples, including those behind branches or wires, so that they can be harvested appropriately. 

Camera movement is exploited to build 3D maps using Structure From Motion (SFM) in~\cite{gene-mola2020a}. SFM is effective where camera motion is smooth and unobscured, but comes at a computational cost and struggles with burring or poor exposure, both common during harvesting. Previous work within the same group examined apple detection using image, depth and radiometric data with Faster-RCNN~\parencite{gene-mola2019b}. This multi-modal data was gathered using a robotic platform at a fixed distance from the trellis and results indicated that early depth fusion alone was not effective, but combining image, intensity and range produced the best detector. 

Bounding boxes can be trivially extracted from image instance semantic segmentation masks, so pixel wise classification is one approach to fruit detection. This technique is considered by~\cite{mccool2016} who trial various local pixel features to perform classification on multi-spectral images, and find local binary features to be most effective. Numerous pixel wise classifiers are applied to specifically detect plums in~\cite{pourdarbani2019} using a variety of manually specified colour space features.

% \cite{yu2019} use Mask-RCNN to detect strawberries for harvesting, a problem exhibiting moderate occlusion and target overlap. Fruit are split into ripe and unripe classes, allowing for single step detection and assessment when harvesting. 

Various sensing modalities for fruit detection, such as hyper-spectral, thermal and stereo vision are explored in~\cite{kapach2012}. Thermal imagery is found to only be useful at certain times of day, while geometry and colour are identified as the strongest features for distinguishing fruit. Similar observations regarding the limitations of thermal imagery use are made in~\cite{bulanon2009}, and by~\cite{gan2018} who also present a novel algorithm for fusing thermal and RGB imagery. 

% \cite{fernandez2014} use multi-spectral and RGB data to train a pixel wise SVM image segmenter, pixels belonging to the fruit class are combined with time of flight camera data to reconstruct fruit geometry. Image sensor metadata, such as camera angle relative to the sun can also improve detection performance~\parencite{bargoti2016}.

Depth data is frequently leveraged for object detection, as in~\cite{tu2020} where a multiscale implementation of Faster-RCNN is improved with the late fusion of RGBD imagery. A Microsoft Kinect V2 camera is used, which requires avoiding direct sunlight and is kept a fixed distance from the fruit trellis. This provides detailed and consistent depth data, intended to be used for fruit counting rather than harvesting. LiDAR sensors have large depth ranges and very high accuracy, but low resolution. \cite{gene-mola2020b} leverage the lighting invariance of LiDAR to detect fruit in point clouds, captured with and without a commercial air blower being applied to the crop. Combinations of data both with, and without, the blower active led to improved single frame detector accuracy, though it was not beneficial for yield prediction. Late fusion of near infra red and RGB imagery is used to detect a range of fruit using Faster-RCNN in~\cite{sa2016} and dataset size is found to have a critical impact on detector performance. 

To the best of the authors' knowledge, this work presents the first example of an eye-in-hand RGBD dataset for object detection gathered during actual tree crop harvesting. Unlike much existing literature, multiple current generation object detector architectures are benchmarked and publicly released, along with day and night datasets.

%EXPERIMENT SECTION
\section{Method}
\begin{figure}[h!]
	\centering
	\includegraphics[width=0.6\columnwidth]{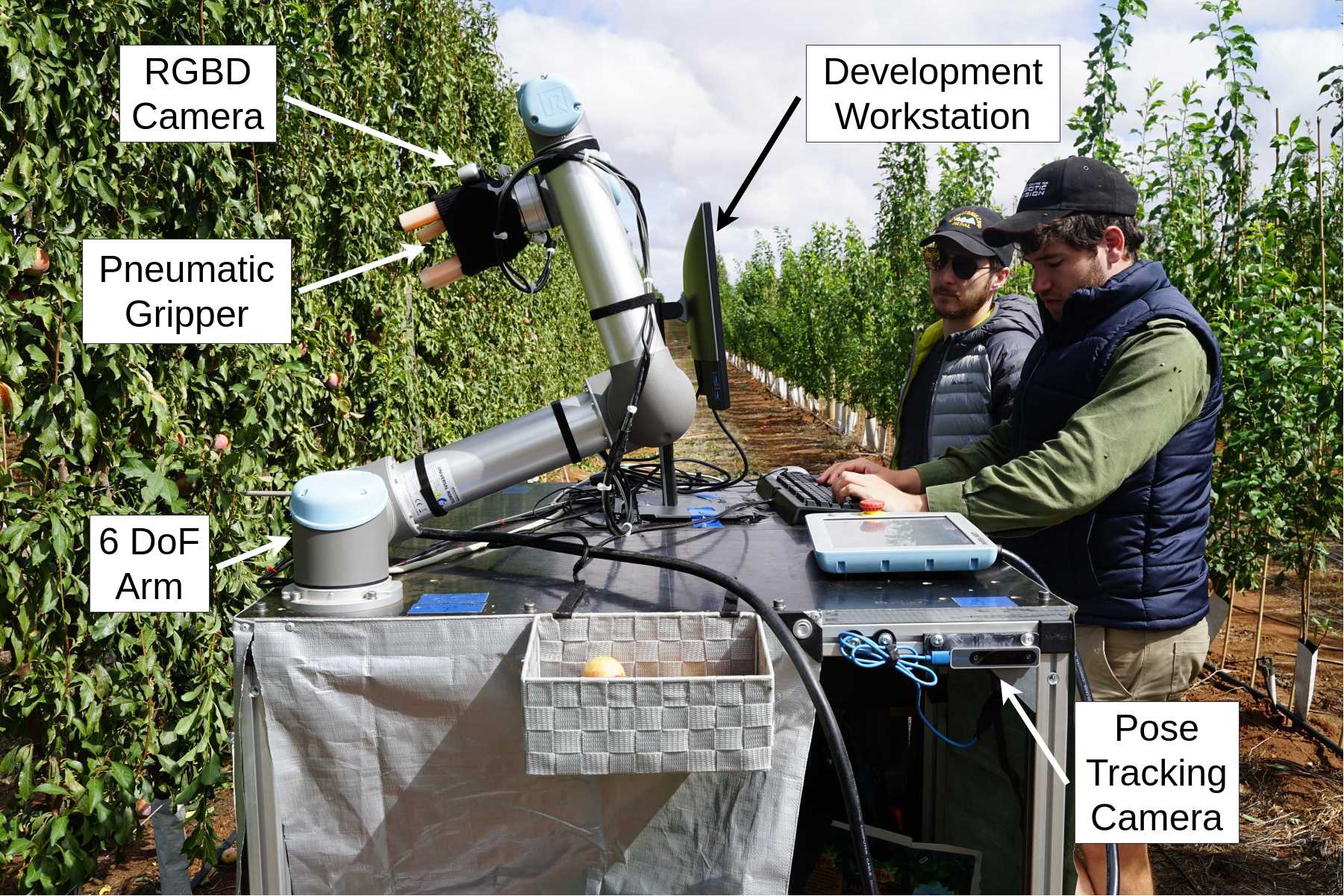}
	\captionof{figure}{The plum harvesting platform including; a 6 DoF robotic arm, RGBD camera, soft pneumatic gripper, mobile base with power and computation, pose tracking camera and development workstation.}
	\label{fig:Platform}
\end{figure}

To test the performance of current generation object detector architectures on a fruit picking task, images are gathered during the trial of a robotic harvesting system, shown in Figure~\ref{fig:Platform}. This consists of an eye-in-hand RGBD camera mounted to a gripper on the end of a 6 Degree of Freedom (DoF) robotic arm. Several hundred images of red plums are gathered during both day and night operations, these are manually labelled to create two datasets. Four deep-learning detectors are trained and evaluated on this data. Each network architecture is trained and tested on the day and night datasets separately. Both pretrained weights for transfer learning from a non-agricultural task, and randomly initialised weights are used with each architecture. Separately to the benchmark tests, RGB and depth fusion is tested with RetinaNet on the day and night datasets. 

% \end{multicols}
% \begin{multicols}{2}

\subsection{Dataset}
\label{sec:dataset}
During an automated harvesting trial of red plums in Victoria, Australia, image feeds from a Realsense D435 RGBD camera were recorded. The crop is grown in a 2D trellis configuration, known as a fruiting wall, with fruit close to branches resulting in difficult harvesting conditions and high rates of target obscuration. The harvesting system travels parallel to this trellis, while the camera and gripper move with five degrees of freedom, three Cartesian plus roll and yaw, to harvest fruit. An embedded version of the YoloV3 model trained on previous data was used to detect harvesting targets with the D435, all detected targets were attempted, resulting in many failed picks in the current dataset. Performance of this embedded model was worse than expected so it is not included in the benchmarking tests. Several hours of sensor data were recorded over the period of one day with direct sunlight, shadowed sunlight and overcast conditions. For night time operation a single diffused floodlight was used, mounted off the arm.

During harvesting the image frames are processed at a rate of 10.5 per second. The camera driver provides on-board depth map alignment to the RGB images and all data is at a resolution of 640x480px. Images for the day dataset were extracted from the camera feed at 0.5 second intervals, then 350 were manually selected to form a representative dataset. Frames that did not contain plums, were excessively blurry, or similar to existing frames were not selected. Colour balance, distance to trellis and number of targets were not considered when selecting frames. This process was repeated for data gathered at night. These two datasets of 350 images each are then manually labelled with bounding boxes around all visible plums and split into train, test and validation subsets of 176, 87 and 87 images respectively. A total of 4449 plums are annotated in the day dataset, and 1402 in the night. Fewer plums were seen at night due to the light source not fully penetrating the canopy.

During harvesting, the camera is positioned approximately 70cm from the trellis for a global view of the trellis area being picked. Three dimensional positions of all detected fruit are filtered using an Extended Kalman Filter (EKF) and then passed to the arm control system which plans a harvest trajectory for each plum. Frames are continuously gathered during the harvesting process and used to improve fruit position estimates within the EKF. Thus the object detector must be robust to both near and far viewpoints of fruit, as seen in the dataset.

\begin{figure}[h!]
	\centering
	\includegraphics[width=0.8\textwidth]{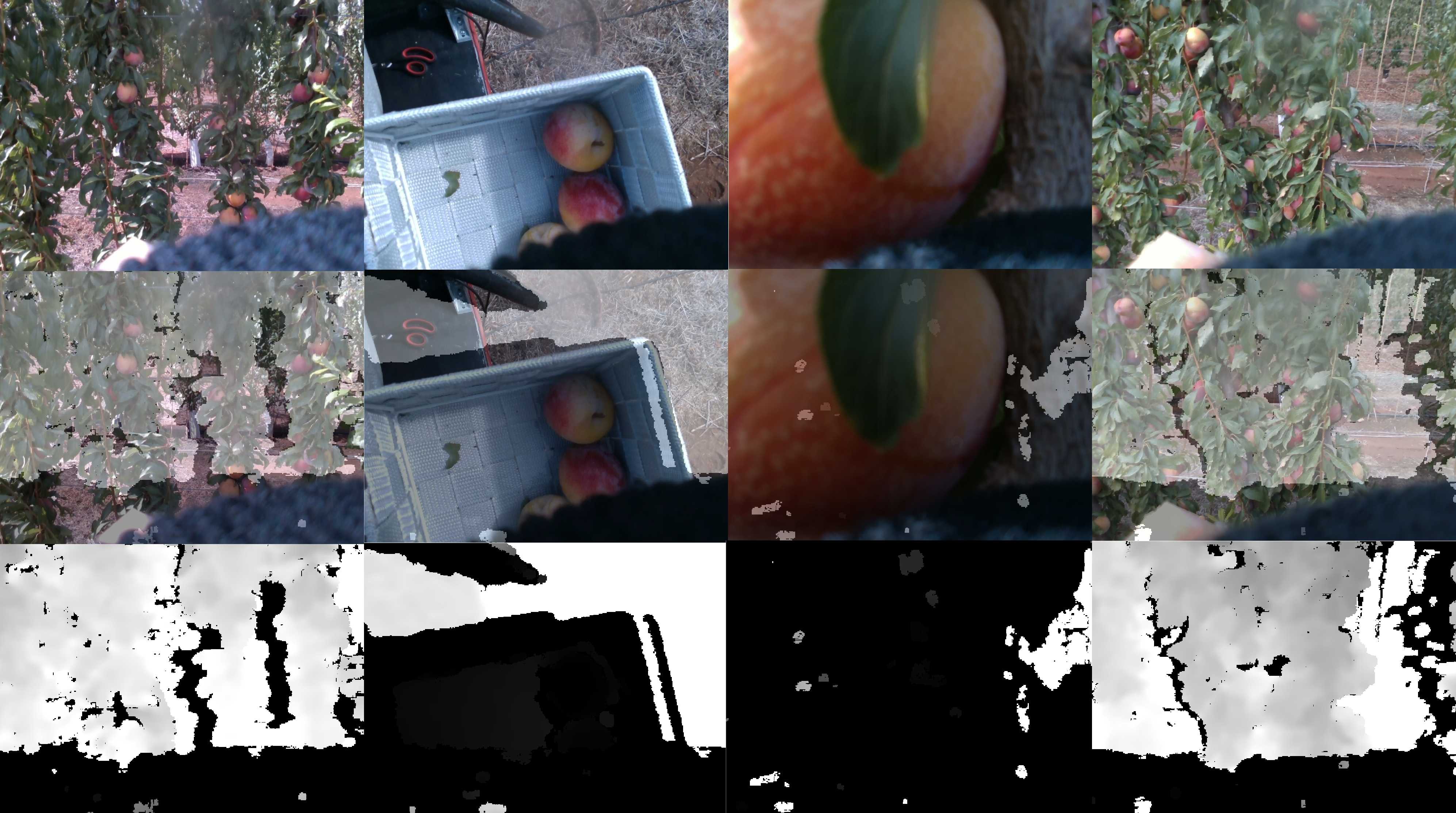}
	\caption{Example RGB, RGBD overlay and depth images from the day dataset. Including effects specific to harvesting motion such as large illumination, object size and obscuration changes. Depth images shown after clipping is applied.}
	\label{fig:datasetDay}
\end{figure}

\begin{figure}[h!]
	\centering
	\includegraphics[width=0.8\columnwidth]{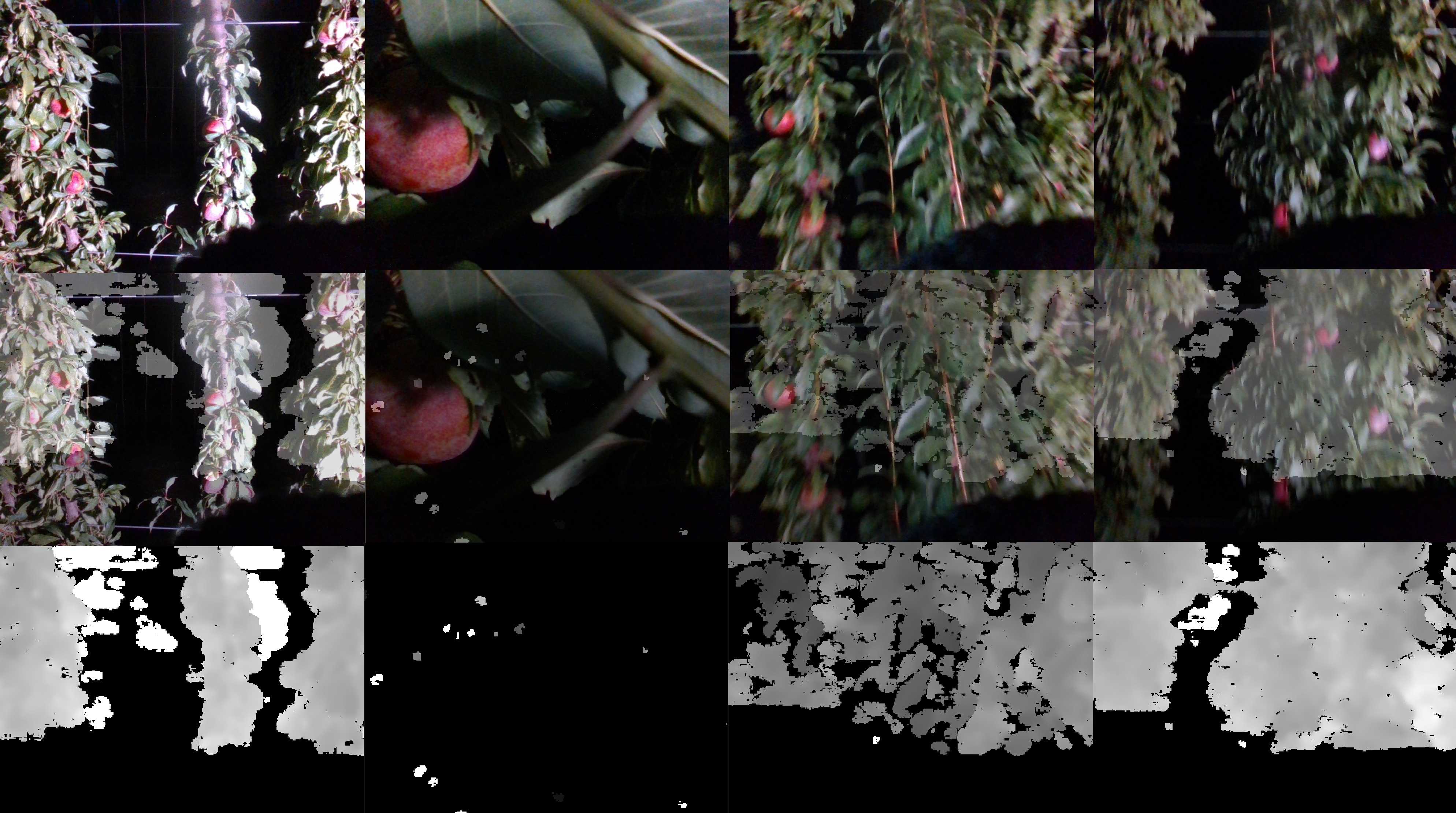}
	\caption{Example RGB, RGBD overlay and depth images from the night dataset. Depth performance is improved at night with better defined object edges and less smoothing effects. Depth images shown after clipping is applied.}
	\label{fig:datasetNight}
\end{figure}

Many of the gathered images exhibit numerous artefacts directly related to the harvesting task. Some of these are caused by the camera motion as it moves from the global pose to harvesting a fruit. This results in a wide range of distances, bounding box sizes and illumination changes, as seen in Figures~\ref{fig:datasetDay} and~\ref{fig:datasetNight}. Most images also include part of the gripper, a design trade-off necessary to minimise the camera and gripper footprint. Exposure and white balance are handled automatically by the camera driver and must be variable to deal with changing light conditions throughout the day. Occasionally this results in extremely mis-exposed or mis-coloured images which the system should be tolerant to and are present in the datasets in small numbers. 

% \end{multicols}
% \begin{multicols}{2}

Depth imagery is required for localising the fruit after they have been detected and is used by the harvesting system. Because this sensor modality is already present, and is increasingly common among automated agricultural platforms, the fusion of RGB and depth data is tested. To create the RGBD dataset each annotated RGB image has a corresponding depth image included as a separate file. Depth data suffers from holes where the range is out of sensor limits, shadowing where a point is obscured for one of the IR stereo pair used to calculate depth, and also from smoothing effects in the depth calculation algorithms. Various methods have been proposed to overcome these limitations, however for simplicity, the depth data is only normalised and clipped before being used in network training. Clipping occurs by setting values below 0.11m, where the camera can return incorrect readings, to be zero and values above 2.5m to be 2.5m. All depth readings are then divided by 2.5m to produce a pixel range from zero to one. At short ranges many depth errors are still present, seen in Figure~\ref{fig:datasetDay} column 3. 

% \end{multicols}
% \begin{multicols}{2}

The datasets are formatted to match the Pascal Visual Object Classes (VOC) 2007 standard~\parencite{everingham2010}. The two RGBD datasets and trained models for these are made available online\footnote{http://data.acfr.usyd.edu.au/Agriculture/PlumDetection/}.

\subsection{RGB Network Architectures}
Four commonly used object detector networks were chosen for evaluation; Faster-RCNN, YoloV3, RetinaNet and CenterNet. The first three use Keras implementations, while CenterNet is in PyTorch. These span a range of target frame rates and all lie on, or close to, the outer edge of the speed-accuracy curve for the standard computer vision dataset Common Objects in COntext (COCO)~\parencite{lin2015}. This can be seen in Table~\ref{table:cocoPerformance}.

Faster RCNN is the only two stage detector tested, an approach which shows improved accuracy over single stage detectors, at the cost of slower inference times~\parencite{ren2017}. YoloV3 is a single stage detector used in many existing works looking at object detection for agriculture, and remains a competitive detector for high frame rate applications. Updated versions of Yolo are also available~\parencite{redmon2018}. 

RetinaNet implements the concept of focal loss which alters the loss function to down-weight the impact of easy negative examples, where there are clearly no objects within the bounding box~\parencite{lin2020b}. CenterNet is the newest and largest network tested~\parencite{duan2019}. A one-stage approach is used to predict heat maps of where bounding box corner and center points lie. 

\begin{table*}
    \caption{The originally published COCO Average Precision metric for each architecture. Additionally, the reported model inference time, although each model uses different GPU hardware so these are only roughly comparable. Faster-RCNN figures are from~\cite{huang2017} who note that model speed is highly dependent on the number of box proposals. Faster-RCNN and RetinaNet resize the input to make the short image edge match the stated value.}
	\begin{center}
	\resizebox{0.7\textwidth}{!}{%
		\begin{tabular}{lcccc}
			\textbf{Network} & \textbf{Backbone} & \textbf{\begin{tabular}[c]{@{}c@{}}Input Size\\ (pix)\end{tabular}} & \textbf{COCO AP} & \textbf{\begin{tabular}[c]{@{}c@{}}Inference Time\\ (ms)\end{tabular}} \\ \hline
			Faster-RCNN & VGG-16 & 600xN & 34.7 & 250 \\
			YoloV3 & DarkNet-53 & 416x416 & 33.0 & 29 \\
			RetinaNet & ResNeXt-101-FPN & 800xN & 40.8 & 198 \\
			CenterNet & Hourglass-104 & 511x511 & 47.0 & 340
		\end{tabular}%

	}
	\label{table:cocoPerformance}
	\end{center}
\end{table*}
% \begin{multicols}{2}

All of the precise training configurations applied to these networks during benchmark testing can also be found at the dataset web page.

\subsection{RGBD Network Architecture}
Two forms of information fusion, early and late, are commonly presented in the literature. Both are tested here using the RetinaNet architecture against an RGB-only baseline. 

\begin{figure}[h!]
	\centering
	\includegraphics[width=1.0\textwidth]{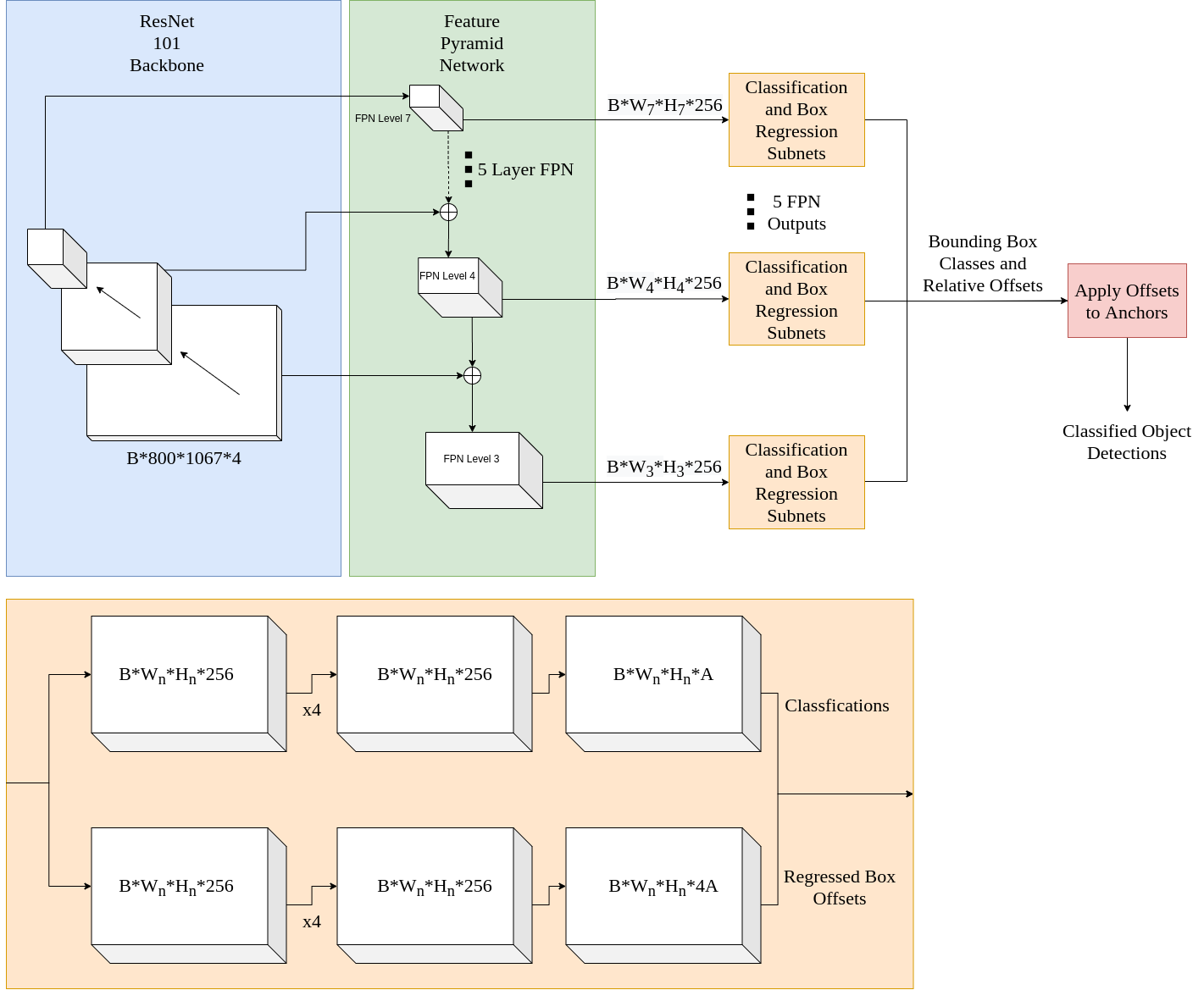}
	\caption{The early RGBD fusion network using the ResNet-101 backbone, adapted from~\cite{lin2020b} and identical to their implementation apart from the input layer shape. B is the batch size, A is the number of anchors, there are a total of 5 FPN levels used which are numbered 3 to 7 to match the above mentioned paper.}
	\label{fig:earlyNetwork}
\end{figure}

Early fusion refers to concatenating the depth information as an additional input channel, in our case this makes the network input a $480 \times 640 \times 4$ tensor, prior to resizing. Early fusion is easily implemented and does not significantly increase computational requirements, but shows mixed results in the literature, often performing worse than RGB alone. Figure~\ref{fig:earlyNetwork} shows the early fusion network.

% \end{multicols}
% \begin{multicols}{2}

Late fusion runs a pair of feature extractor backbones and FPNs, on the RGB and depth data in parallel. At each of the 5 FPN scales, features from the RGB and depth FPN outputs are channel wise stacked before being passed through a 1x1 convolution which performs pooling over the RGB and depth feature maps. This reduces the FPN channels to 256 so the classification and regression subnetworks are identical to the RGB-only case. The overall network size is slightly less than doubled. Addition of an extra backbone creates more informative features which can be learned specifically for the depth modality, at the cost of additional complexity and execution time. Depth data has a meaningful absolute value and is pre-normalised to a fixed range, so batch normalisation layers are removed from the depth backbone. 

% \end{multicols}
\begin{figure}[h!]
	\centering
	\includegraphics[width=1.0\textwidth]{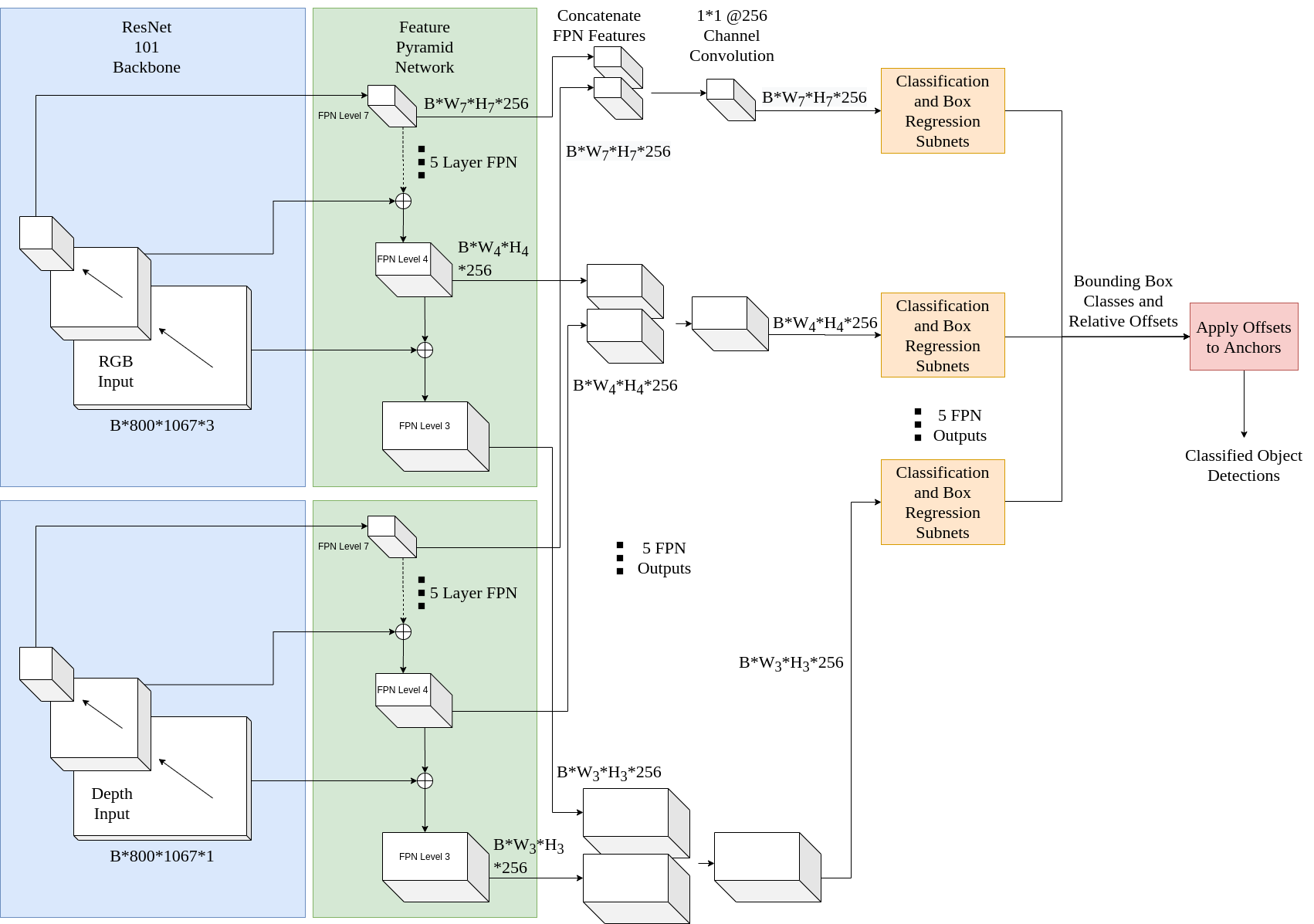}
	\caption{The late RGBD fusion network, adapted from~\cite{lin2020b}. The backbone and FPN is duplicated, output features from this are channel wise stacked and passed through a $1\times 1$ convolution to reduce the channels back to 256. The same two subnets as Figure~\ref{fig:earlyNetwork} are then applied.}
	\label{fig:lateNetwork}
\end{figure}
% \begin{multicols}{2}

\subsection{Training \& Testing Details}
All object detection networks can be improved through careful hyperparameter tuning. To provide a fair comparison, reduce evaluation time and because many application areas lack the resources for extensive network tuning, each architecture is trained using the default parameters provided by the authors. For YoloV3 and CenterNet these were found using the COCO dataset, while Faster-RCNN and RetinaNet were primarily developed using Pascal VOC. 

Modifying the default anchor proposals to better suit plums was found to be counterproductive, so default anchor sizes are used for each network. All architectures do some form of data augmentation by default, specifics of which can be found at public link provided. Networks are trained until the validation loss plateaus. All inference time results were achieved using an Nvidia GTX 1080Ti and training batch sizes are set to the maximum that can fit on this GPU.

Transfer learning refers to using weights from an already trained model as the starting point for training on the day and night plums datasets. All networks are tested both with and without transfer learning.

All models are modified to produce a Pascal VOC format results file for the test set, which are then processed using the official VOC2007 Matlab development kit~\parencite{everingham2010}. Evaluation is done by plotting the Precision-Recall (PR) curve and reporting the official Average Precision (AP) metric with a bounding box Intersection Over Union (IOU) threshold of 0.5.

Depth data is absolute in nature and relating scene geometry to image data requires the camera focal length. So to preserve correlated features between RGB and depth inputs, a fixed focal length should be used. This prevents the use of image re-scaling and the augmentations, such as cropping, translation and rotation, which rely on it. 

For all six RGBD tests, image augmentation is disabled and transfer learning is applied using ImageNet weights. The dual backbones were found to make late fusion training unstable so a two step process is required to effectively train this network. First the depth backbone is frozen and the RGB ResNet, FPN and subnet modules are trained, then all layers are then unfrozen and the depth backbone is also trained. 

%RESULTS SECTION
\section{Results}
\begin{figure}
	\centering
	\begin{subfigure}{.5\textwidth}
		\centering
		\includegraphics[width=1\columnwidth]{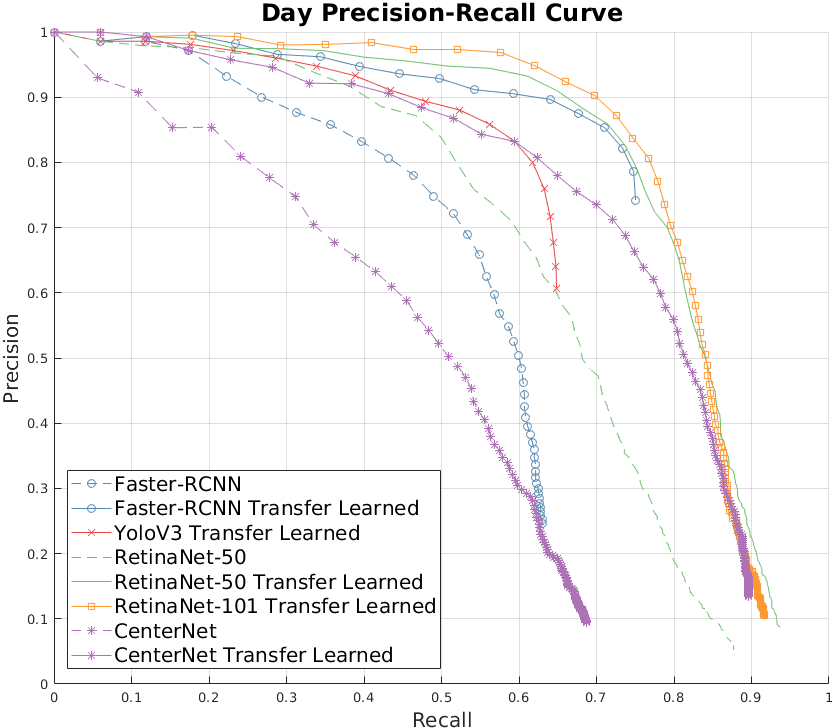}
	\end{subfigure}%
	\begin{subfigure}{.5\textwidth}
		\centering
		\includegraphics[width=1\columnwidth]{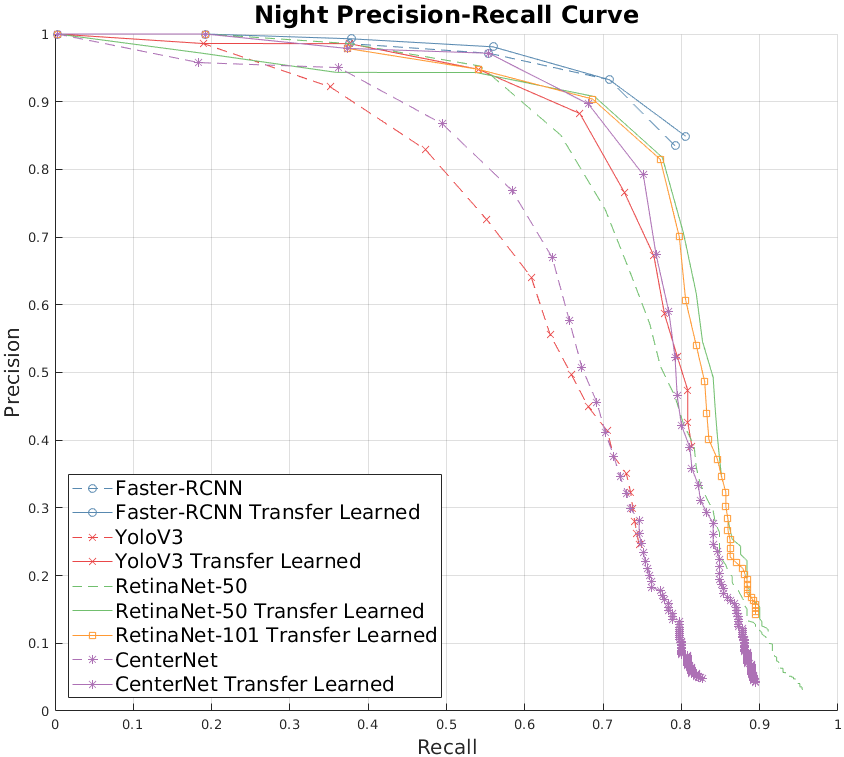}
	\end{subfigure}
	\caption{The PR curve for each architecture on the day time (left) and night time (right) datasets.}
	\label{fig:PR_RGB}
\end{figure}

Each RGB network is tested against the day and night dataset separately, both with and without transfer learning. The impact of depth fusion is assessed using the RetinaNet architecture. All tested configurations are summarised in Table~\ref{table:results} with PR curves for each dataset shown in Figure~\ref{fig:PR_RGB}.

Some training runs were unstable, resulting in no validation set AP increase during training. Each unstable training configuration was tested three times and in all cases the three runs failed. No training runs failed where an AP value is reported.

% \end{multicols}
\begin{table}
	\begin{center}
	\caption{Results for each network on the day and night datasets. AP is calculated using the VOC2007 development kit, mean inference time is per image, not including network loading time.}
		\resizebox{0.8\textwidth}{!}{%
			\begin{tabular}{lllccc}
				\multicolumn{1}{c}{\textbf{Architecture}} & \multicolumn{1}{c}{\textbf{Backbone}} & \multicolumn{1}{c}{\textbf{Configuration}} & \textbf{\begin{tabular}[c]{@{}c@{}}Day AP \\ @0.5 IOU\end{tabular}} & \textbf{\begin{tabular}[c]{@{}c@{}}Night AP \\ @0.5 IOU\end{tabular}} & \textbf{\begin{tabular}[c]{@{}c@{}}Mean Inference\\ Time (ms)\end{tabular}} \\ \hline
				\multirow{2}{*}{Faster-RCNN} & VGG-16 & Transfer Learned & 0.691 & \textbf{0.795} & \multirow{2}{*}{128} \\
				& VGG-16 & Random Weights & 0.537 & 0.788 &  \\
				&  &  &  &  &  \\
				\multirow{2}{*}{YoloV3} & DarkNet-53 & Transfer Learned & 0.597 & 0.746 & \multirow{2}{*}{\textbf{56}} \\
				& DarkNet-53 & Random Weights & Unstable & 0.608 &  \\
				&  &  &  &  &  \\
				\multirow{2}{*}{RetinaNet} & ResNet-50 & Transfer Learned & 0.781 & 0.778 & \multirow{2}{*}{72} \\
				& ResNet-50 & Random Weights & 0.639 & 0.744 &  \\
				&  &  &  &  &  \\
				\multirow{2}{*}{RetinaNet} & ResNet-101 & Transfer Learned & \textbf{0.787} & 0.767 & \multirow{2}{*}{102} \\
				& ResNet-101 & Random Weights & Unstable & Unstable &  \\
				&  &  &  &  &  \\
				\multirow{2}{*}{CenterNet} & Hourglass-104 & Transfer Learned & 0.709 & 0.746 & \multirow{2}{*}{276} \\
				& Hourglass-104 & Random Weights & 0.456 & 0.632 &  \\
				&  &  &  &  &  \\ \hline
				\multirow{3}{*}{\begin{tabular}[c]{@{}l@{}}Retinanet \\ RGBD\end{tabular}} & ResNet-101 & Early Depth Fusion & 0.608 & 0.732 & 109 \\
				& ResNet-101 & Late Depth Fusion & \textbf{0.745} & \textbf{0.781} & 143 \\
				& ResNet-101 & RGB Baseline & 0.730 & \textbf{0.781} & \textbf{99}
			\end{tabular}%
		}
		\label{table:results}
	\end{center}
\end{table}
% \begin{multicols}{2}

%For figures side by side
% \begin{figure}[h]
% 	\centering
% 	\begin{minipage}{.49\textwidth}
% 		\centering
% 		\includegraphics[width=1.0\columnwidth]{Figs/DayPR.png}
% 		\caption{The PR curve for each architecture on the day time dataset.}
% 		\label{fig:PRDay}
% 	\end{minipage} \hfill
% 	\begin{minipage}{.49\textwidth}
% 		\centering
% 		\includegraphics[width=1.0\columnwidth]{Figs/NightPR.png}
% 		\caption{The PR curve for each architecture on the night time dataset.}
% 		\label{fig:PRNight}
% 	\end{minipage}
% \end{figure}

% \begin{figure}[h!]
% 	\centering
% 	\includegraphics[width=1.0\columnwidth]{Figs/DayPR.png}
% 	\captionof{figure}{The PR curve for each architecture on the day time dataset.}
% 	\label{fig:PR_Day}
% \end{figure}

% \begin{figure}[h!]
% 	\centering
% 	\includegraphics[width=1.0\columnwidth]{Figs/NightPR.png}
% 	\captionof{figure}{The PR curve for each architecture on the night time dataset.}
% 	\label{fig:PR_Night}
% \end{figure}

%For figures side by side one caption

\subsection{RGB Only}
Over the four baseline networks tested, RetinaNet with ResNet-101 achieved the highest AP on the day dataset while Faster-RCNN performed best on the night data. Transfer learned networks were much more accurate than those trained from scratch, while also taking less time to train. 

YoloV3 was the fastest network by a significant margin, although with lower than average accuracy. RetinaNet with ResNet-50 provides a good speed-accuracy trade off for most applications. Data augmentation using the RetinaNet default methods was effective, shown by the difference between the RGB baseline from the RGBD tests and the ResNet-101 transfer learned results. 

Faster-RCNN, YoloV3 and CenterNet all performed better on the night dataset. Fixed lighting conditions and fewer visible but obscured fruit should make this an easier detection task, though there are less training instances available.

\subsection{RGBD Fusion} 
\begin{figure}[h!]
	\centering
	\includegraphics[width=0.49\columnwidth]{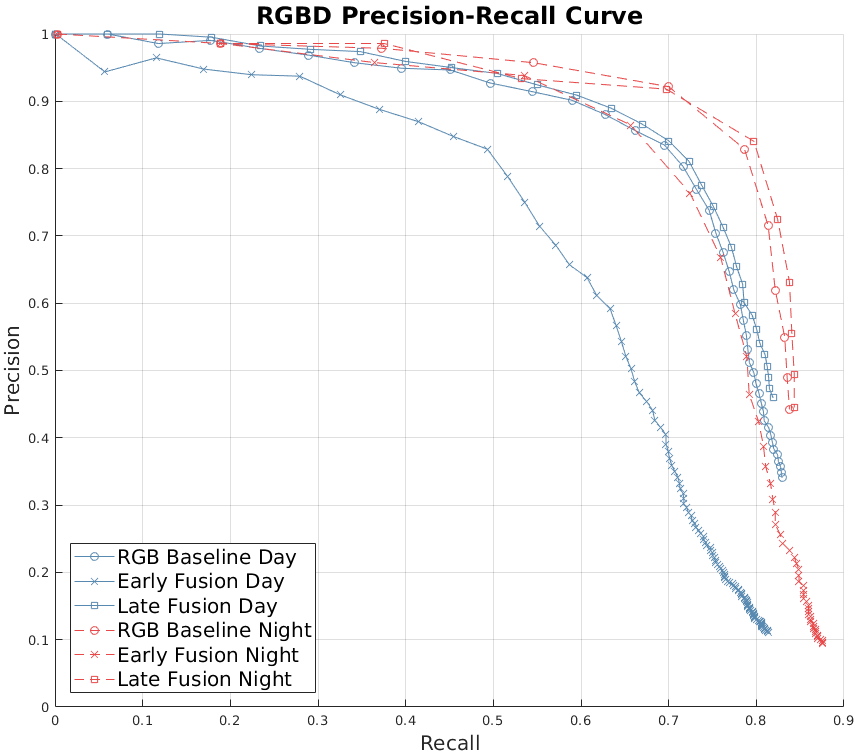}
	\captionof{figure}{The PR curve for the three RGBD fusion approaches on both datasets.}
	\label{fig:PR_RGBD}
\end{figure}

Precision-recall curves for the RGBD tests are plotted in Figure~\ref{fig:PR_RGBD}. Early fusion performed worse than RGB alone, even with other factors such as data augmentation, being equal. Late fusion slightly outperformed the baseline on both day and night data. Doubling of the network backbone produced only a 31\% increase in inference time. Many operations such as image pre-processing and bounding box non-maxima suppression are not dependent on network size. 

% \end{multicols}
% \begin{multicols}{2}

\section{Discussion}
Differences between performance reported on the COCO dataset, and the two datasets tested, were surprising. CenterNet performed poorly on both plums tasks, despite having the highest stated COCO accuracy, it was also the slowest. RetinaNet was effective for day time detection, with augmentation, transfer learning and backbone size all playing a role in overall performance. 

Applying transfer learning had an overall larger impact than architecture selection and is essential when using small datasets, as in many agricultural applications. Conversely, using the ResNet-101 backbone significantly increased RetinaNet processing time, for only a small benefit in precision. Design decisions such as these can play a more essential role than architecture choice, and should be carefully considered. Additional factors, such as dataset and batch size, are not investigated in this work but typically also have an impact on accuracy. 

Faster-RCNN outperformed both more modern and slower networks on the night dataset. The reason for this is unknown, and is in contrast to the daytime performance. Fortuitous hyperparameter defaults may be a contributing factor, though properly exploring these for all 10 RGB configurations is not feasible. This result highlights the importance of testing a range of network types on application specific data, such as harvesting under controlled lighting.

Early data fusion was counterproductive for these datasets. Although late data fusion was effective, the gains from this method were less than that provided by data augmentation, and expanding the RGB dataset size would likely be significantly more useful than incorporating depth information. Additionally, longer ResNet backbones exist and typically show a small AP improvement over ResNet101. So the additional network capacity introduced by the late fusion approach may be better used as a longer RGB only backbone.

Predicting the full extent of partially obscured fruit is a requirement of the harvesting system and is well met by all of the tested networks. No accuracy metric is ideal for all use cases, and the 0.5 IOU threshold is an arbitrary assessment point commonly used in computer vision. Other metrics may be more suitable for tasks such as harvesting, where the IOU threshold required for a successful pick can often be estimated. 

The overall AP values for both datasets may be sufficient for commercial fruit harvesting, but definitely show room for improvement. Lack of standardised datasets for testing has been an issue when trying to measure improvement in both robotic harvesting and fruit detection tasks. We hope that releasing this dataset and suite of trained models goes some way towards addressing this for plums. 

%CONCLUSIONS SECTION
\section{Conclusions}
In this work two datasets gathered during a robotic harvesting trial on 2D trellis plums are presented, and four deep learning object detection architectures are benchmarked on these. The fusion of depth information was trialled and found to be marginally effective for late fusion, though data augmentation provides a larger performance boost. On the day time dataset RetinaNet was the most accurate, while Faster-RCNN showed the best average precision for the night time data. 

Relative network performance differed significantly to that published for the COCO dataset, which is commonly used when making design decisions for applications in agriculture. So the public availability of a wide variety of application specific datasets, such as tree fruit harvesting, is important to future progress. 

During the next harvesting season the limited dataset size can be addressed by expanding the amount of plum picking data and testing on additional fruit types. Future investigations into accuracy metrics specific to tree crop harvesting, multi-view detection and multi-spectral imaging are also planned.

\newpage
\printbibliography

\newpage

%\theendnotes FOR FOOTNOTES AT END

% \end{multicols}

\end{document}